\documentclass[10pt,twocolumn,letterpaper]{article}
\usepackage{wacv}
\wacvfinalcopy % TODO: comment out for submission
\usepackage{times}
\usepackage{epsfig}
\usepackage{graphicx}
\usepackage{amsmath}
\usepackage{amssymb}
\usepackage{caption}
\usepackage{cuted}% for environment `strip`
\usepackage{bm}
\def\wacvPaperID{1117} % Enter the WACV Paper ID here

\ifwacvfinal
\usepackage[breaklinks=true,bookmarks=false]{hyperref}
\else
\usepackage[pagebackref=true,breaklinks=true,colorlinks,bookmarks=false]{hyperref}
\fi

\usepackage[noabbrev]{cleveref}

\begin{document}

%%%%%%%%% TITLE
\title{Hybrid-S2S: Video Object Segmentation with Recurrent Networks and Correspondence Matching}

\author{Fatemeh Azimi\textsuperscript{1,2}\qquad Stanislav Frolov\textsuperscript{1,2}\qquad Federico Raue\textsuperscript{2}\qquad J\"orn Hees\textsuperscript{2}\qquad Andreas Dengel\textsuperscript{1,2}\\
\textsuperscript{1}TU Kaiserslautern, Germany\\ 
\textsuperscript{2}DFKI GmbH, Germany\\ 
\url{firstname.lastname@dfki.de}
}

\maketitle
% https://tex.stackexchange.com/questions/470493/span-image-over-two-columns-without-perturbing-the-text-flow
\begin{strip}
    \centering
    \includegraphics[width=\textwidth]{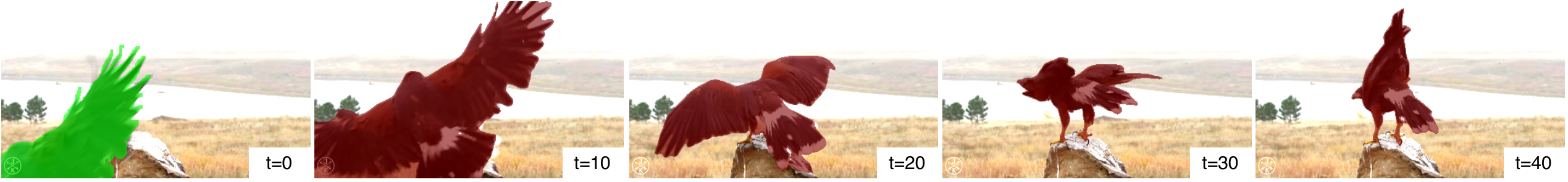}
    \captionof{figure}{In one-shot video object segmentation, the object mask at $t=0$ is available. The model is trained to track an object of interest throughout the rest of the video sequence. In this figure, the green mask in the first frame is the ground-truth. The remaining segmentation masks~(red) are the output of our HS2S model.}
    \label{fig:teaser}
\end{strip}

%\thispagestyle{empty}
%%%%%%%%%%%%%%%%%%%%%%%%%%%%%%%%%%%%%%%%%%%%%%%%%%%%
%
%       
%       Abstract    
%
%
%%%%%%%%%%%%%%%%%%%%%%%%%%%%%%%%%%%%%%%%%%%%%%%%%%%%
\begin{abstract}
%\vspace{-5mm}
One-shot Video Object Segmentation~(VOS) is the task of pixel-wise tracking an object of interest within a video sequence, where the segmentation mask of the first frame is given at inference time.
In recent years, Recurrent Neural Networks~(RNNs) have been widely used for VOS tasks, but they often suffer from limitations such as drift and error propagation. 
In this work, we study an RNN-based architecture and address some of these issues by proposing a hybrid sequence-to-sequence architecture named HS2S, utilizing a dual mask propagation strategy that allows incorporating the information obtained from correspondence matching.
Our experiments show that augmenting the RNN with correspondence matching is a highly effective solution to reduce the drift problem.
The additional information helps the model to predict more accurate masks and makes it robust against error propagation.
We evaluate our HS2S model on the DAVIS2017 dataset as well as Youtube-VOS.
On the latter, we achieve an improvement of 11.2pp in the overall segmentation accuracy over RNN-based state-of-the-art methods in VOS. 
We analyze our model's behavior in challenging cases such as occlusion and long sequences and show that our hybrid architecture significantly enhances the segmentation quality in these difficult scenarios.
\end{abstract}
%%%%%%%%%%%%%%%%%%%%%%%%%%%%%%%%%%%%%%%%%%%%%%%%%%%%
%
%       
%       Introduction    
%
%
%%%%%%%%%%%%%%%%%%%%%%%%%%%%%%%%%%%%%%%%%%%%%%%%%%%%
%\vspace{-5.5mm}
\section{Introduction}
One-shot Video Object Segmentation~(VOS) aims to segment an object of interest in a video sequence, where the object mask in the first frame is provided.
The objective of this task is to track a target object in a pixel-wise manner.
It has various applications such as robotics, autonomous driving, and video editing to name a few.
VOS is a challenging task, and generating quality segmentation masks requires addressing inevitable real-world difficulties such as unconstrained camera motion, occlusion, fast motion, and motion blur as well as handling objects with different sizes.

VOS has been extensively studied in the Computer Vision community with several works based on classical techniques such as energy minimization and utilizing superpixels~\cite{chang2013video,marki2016bilateral,grundmann2010efficient}.
However, learning-based methods~\cite{perazzi2017learning,Man+18b} have proved to be more successful by significantly surpassing the traditional approaches.

Amongst the wide variety of the suggested learning-based methods, some works approach the problem by processing the frames independently and learning an object model~\cite{perazzi2017learning,Man+18b}, while others utilize temporal information~\cite{xu2018youtub,ventura2019rvos}.
Tokmakov \etal~\cite{tokmakov2017learning} propose utilizing optical flow to propagate the object mask throughout the sequence and make use of the motion cues as well as the spatial information.
However, flow-based models need an additional component for flow estimation~\cite{ilg2017flownet}, which is usually trained separately, and the performance of the whole system is dependent on the accuracy of this module.
With the same motivation of using temporal data,~\cite{xu2018youtub,ventura2019rvos,azimi2020revisiting} utilize Recurrent Neural Networks~(RNNs) to track the target object in a temporally consistent way.
These models are trained end-to-end and rely on learning the spatio-temporal features to track the object and to propagate the object mask across time.
A disadvantage of this category is the performance drop in longer sequences caused by drift and error propagation in the RNN.

\begin{figure}[t]
\centering
  \includegraphics[width=0.9\linewidth]{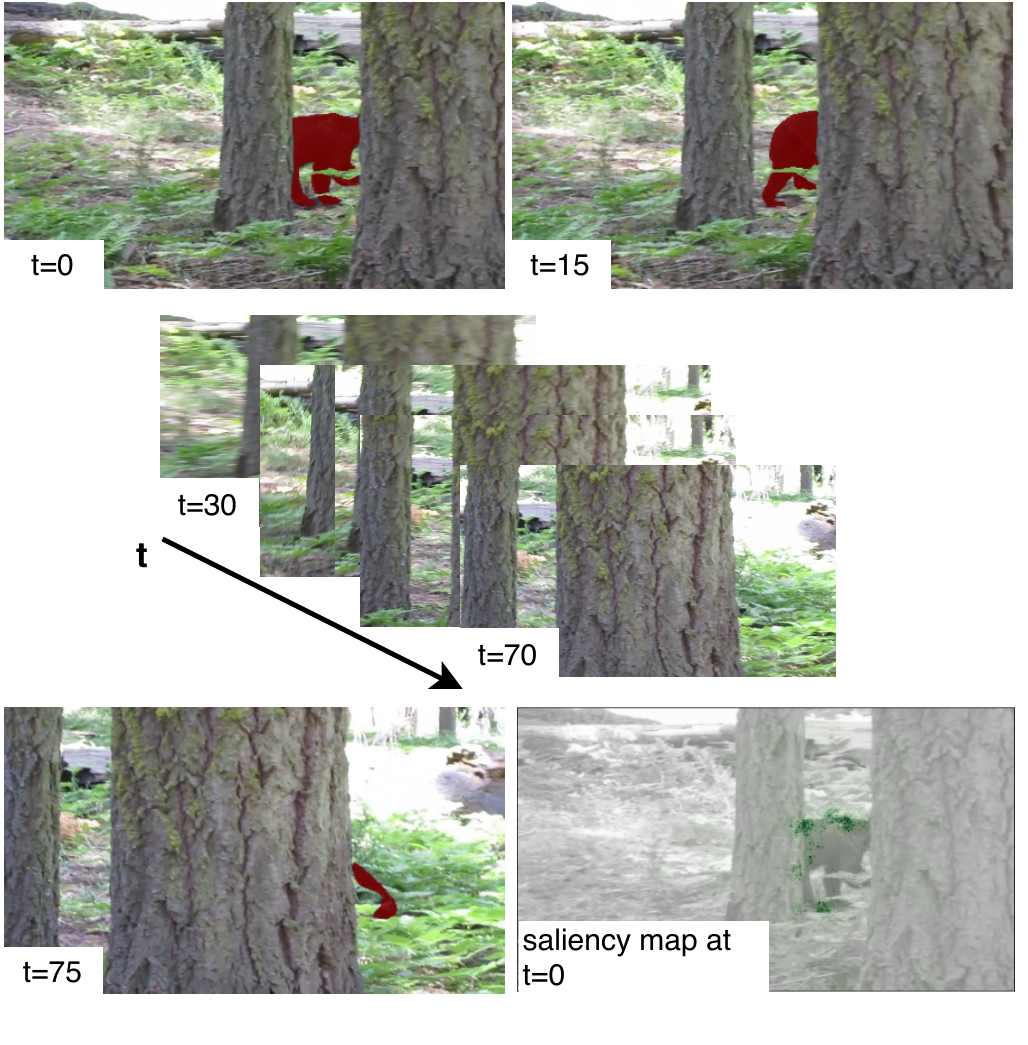}
   \caption{This figure indicates how utilizing the first frame as the reference can help the model recover from occlusion. Here, the object of interest is a bear overlaid with the red mask, which is absent from the middle row frames (from $t=30$ to $t=70$).
   We observe that the model can detect the animal after it appears again, and by looking at the saliency map of the first frame, we note that the model has correctly captured the correspondence between the bear in the first frame and the frame right after the occlusion.
   }
\label{fig:occlusion}
\end{figure}

In this work, we study S2S~\cite{xu2018youtub}, a common RNN-based model for VOS due to the effectiveness of RNNs in utilizing the spatio-temporal features and providing a motion model of the target object, resulting in good segmentation accuracy.
Inspired by~\cite{faktor2014video,wug2018fast,yang2019anchor}, we propose a dual propagation strategy by augmenting the spatio-temporal features obtained from the RNN with correspondence matching to reduce the impact of drift.
Utilizing the features obtained from similarity matching provides a robust measurement for segmentation, improves the segmentation quality, and reduces the error propagation.
This aspect is especially beneficial for the model in long sequences where the RNN performance declines.
Additionally, we integrate the first frame features into the model throughout the whole sequence as a reliable source of information~\cite{ebert2017self,wug2018fast,oh2019video,yang2019anchor}.
By employing these reference features, the model can better handle challenging scenarios such as occlusion~\cite{ebert2017self}, since, by definition, the object is present in the first frame.
\autoref{fig:occlusion} shows an illustration of how correspondence matching together with utilizing the first frame can be helpful in better handling the occlusion.
We hypothesize that the RNN also plays a complementary role in correspondence matching.
Imagine a scenario where multiple instances of similar objects are present in the scene;
in this case, the spatio-temporal model learned by the RNN can act as a location prior and aid the model to distinguish between the target object and the other similar instances.

We evaluate our hybrid sequence-to-sequence~(HS2S) method on the Youtube-VOS~\cite{xu2018youtub} and DAVIS2017~\cite{pont20172017} datasets and demonstrate that our model significantly improves the independent RNN-based models' segmentation quality \cite{xu2018youtub,ventura2019rvos}.
%%%%%%%%%%%%%%%%%%%%%%%%%%%%%%%%%%%%%%%%%%%%%%%%%%%%
%
%       
%       Related Work    
%
%
%%%%%%%%%%%%%%%%%%%%%%%%%%%%%%%%%%%%%%%%%%%%%%%%%%%%
\section{Related Work}
A large body of research in Computer Vision literature has studied VOS during the last decade.
%VOS has a long history in Computer Vision with a large body of research dedicated to this task. 
The classical methods for solving VOS were mainly based on energy minimization~\cite{brox2010object,faktor2014video,papazoglou2013fast,shankar2015video}. 
Brox \etal~\cite{brox2010object} propose a model based on motion clustering and segment the moving object via the analysis of the point trajectories throughout the video.
They also use motion cues to distinguish foreground from background.
% important, ideas about similarity matching
Faktor \etal~\cite{faktor2014video}
present a method based on consensus voting.
They extract the superpixels in each frame, and by computing the similarity of the superpixel descriptors, then use the nearest neighbor method to cluster the most similar superpixels together in a segmentation mask.
\cite{jain2014supervoxel} addresses the problem of fast motion and appearance change in the video by extending the idea of using superpixels to using super-voxels (adding the time dimension) and taking into account the long-range temporal connection during the object movement.

Since the advent of Deep Learning~\cite{krizhevsky2012imagenet}, the Computer Vision community has witnessed a significant progress in the accuracy of VOS methods~\cite{Man+18b,perazzi2017learning,tokmakov2017learning}.
The success of learning based methods can largely be accounted to progress made in learning algorithms~\cite{krizhevsky2012imagenet, he2016deep} and the availability of large-scale VOS datasets such as Youtube-VOS~\cite{xu2018youtub}. 

% online offline
In one-shot VOS, there exist two training schemes, namely offline and online training. Offline training is the standard training phase in learning-based techniques.
As the segmentation mask of the first object appearance is available at test time, online training refers to further fine-tuning the model on this mask with extensive data augmentation.
This additional step considerably improves the segmentation quality at the expense of slower inference.

Considering offline training, one can divide the proposed solutions into multiple categories.
Some methods focus on learning the object masks using only the frame-wise data~\cite{perazzi2017learning, Man+18b}.
In~\cite{Man+18b}, authors extended a VGG-based architecture designed for retinal image understanding~\cite{maninis2016deep} for VOS.
They start with the pre-trained weights on ImageNet~\cite{deng2009imagenet}, and then further train the \textit{parent network} on a specialized VOS dataset~\cite{Perazzi2016}.
This model relies on online training and boundary snapping for achieving good performance.
\cite{voigtlaender2017online} further improves this method by employing online adaption to handle drastic changes in the object's appearance.
Perazzi \etal~\cite{perazzi2017learning} provide a solution based on guided instance segmentation.
They utilize a DeepLab architecture~\cite{chen2017deeplab} and modify the network to accept the previous segmentation mask as an additional input.
Therefore, a rough guidance signal is provided to the model to mark the approximate location where the object of interest lies.
%They show that different guiding signals, such as optical flow or bounding box, achieve similar results.
Yang \etal~\cite{yang2018efficient} take a meta-learning approach and train an additional modulator network that adjusts the middle layers of a generic segmentation network to capture the appearance of the target object.

In~\cite{wug2018fast} a Siamese architecture is used to segment the object based on its similarity to the mask template in the first frame.
Similarly, \cite{yang2019anchor} proposes a zero-shot VOS model, where the object mask at every time step is detected based on the similarity of the current frame to the anchor frames~(first frame and the frame at $t-1$). 
Following this idea, \cite{johnander2019generative} suggests a generative approach for segmenting the target object, introducing an appearance module to learn the probabilistic model of the background and the foreground object.
In~\cite{zhang2020transductive}, the authors develop a model that propagates the segmentation mask based on an affinity in the embedding space.
They propose to model the local dependencies via using motion and spatial priors and the global dependencies based on the visual appearance learned by a convolutional network.
Although these methods obtain good performance on the standard benchmarks~\cite{Perazzi2016,pont20172017}, they do not utilize temporal information and motion cues.

%RCNN
Another line of work relies on region proposal techniques such as \cite{he2017mask}.
For example, \cite{luiten2018premvos} takes a multi-step approach, in which they first generate the region proposals and then refine and merge promising regions to produce the final mask.
Furthermore, they use optical flow to maintain the temporal consistency.
In~\cite{li2017video}, an additional re-identification method based on template-matching is used.
This way, the model can recapture objects lost at some point in the sequence.
These methods are quite complex in architecture design and relatively slow at inference time.

%memory and optical flow
A different group of methods focus on utilizing a memory module to process motion and compute spatio-temporal features.
In order to obtain temporally consistent segmentation masks,~\cite{xu2018youtub, azimi2020revisiting,tokmakov2017learning,ventura2019rvos} employ a ConvLSTM~\cite{xingjian2015convolutional} (or ConvGRU) memory module while~\cite{oh2019video} resorts to using an external memory to process the space-time information.

In this work, we build on top of the S2S~\cite{xu2018youtub} architecture, which is an RNN-based method, on account of exploiting the spatio-temporal features, good performance, and the simple architecture. 

We study some of this model's shortcomings stemming from the finite memory and error propagation in RNNs. 
To address these limitations, we propose a hybrid design that combines the spatio-temporal features from the RNN with similarity matching information, as it will be elaborated in the next section.
Unlike~\cite{oh2019video}, our model does not require any form of external memory.
This is advantageous since using external memory results in additional constraints in the inference phase (e.g. memory overflow for long video sequences).
%%%%%%%%%%%%%%%%%%%%%%%%%%%%%%%%%%%%%%%%%%%%%%%%%%%%
%
%       
%       Method    
%
%
%%%%%%%%%%%%%%%%%%%%%%%%%%%%%%%%%%%%%%%%%%%%%%%%%%%%
\section{Method}
In this section, we explain our hybrid architecture for VOS.
We build on top of the S2S model~\cite{xu2018youtub}, which is an RNN-based architecture and employ a dual mask propagation strategy that utilizes the spatio-temporal features from the RNN as well as correspondence matching to propagate the mask from time step $t-1$ to $t$. 
Moreover, we integrate the features from the first frame as a reference throughout the sequence.
%transferred to later, as suggested in \cite{ebert2017self, wug2018fast, yang2019anchor}.

\begin{figure*}[t!]
    \centering
    % \captionsetup{width=15cm}
    \includegraphics[width=16cm]{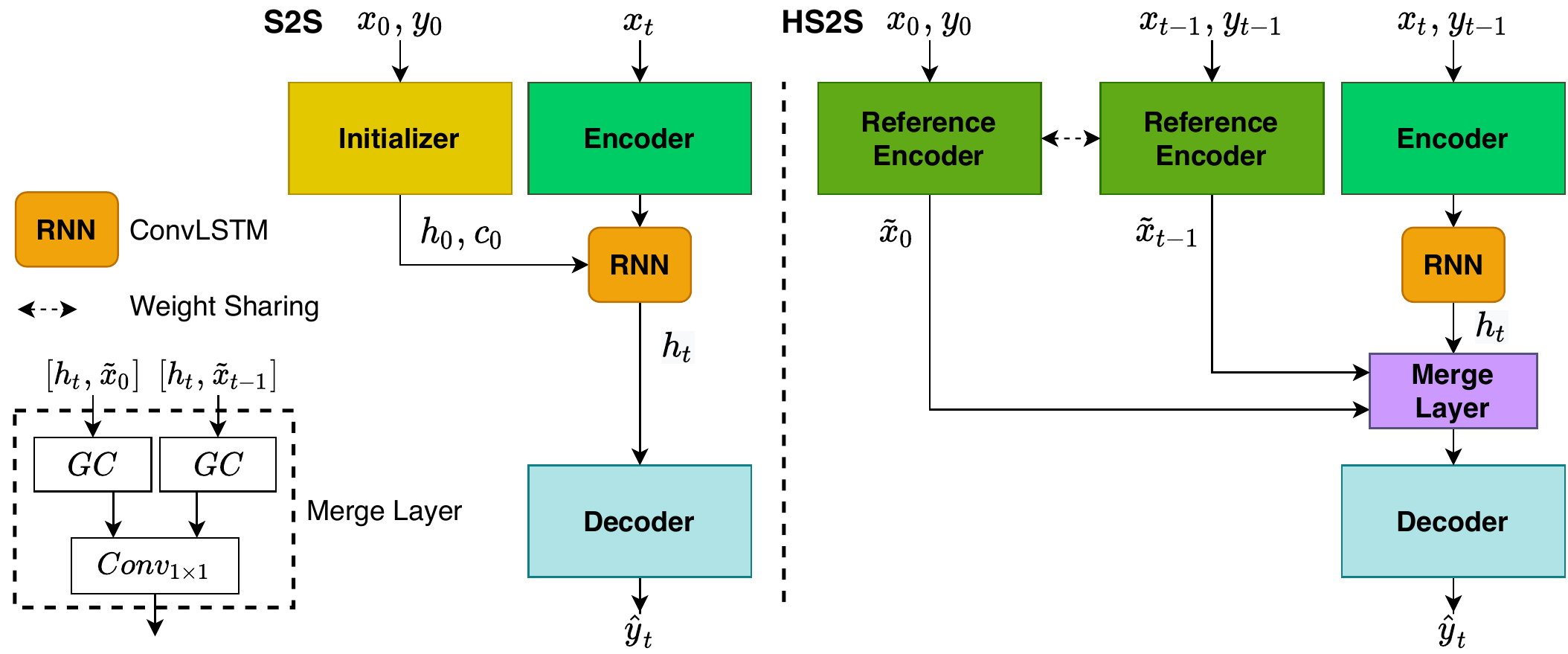}
    \caption{In this figure, we depict the overall architecture of S2S \cite{xu2018youtub}~(\Cref{eq:1,eq:2,eq:3,eq:4}) and our HS2S method~(\cref{eq:5,eq:x0,eq:xt_1,eq:6,eq:7,eq:8}). 
    In HS2S, we initialize the RNN hidden states~($h_{0}$ and $c_{0}$) with \textit{zeros}, instead of using the initializer network.
    We keep track of the target object by feeding the previous segmentation mask~($y_{t-1}$) to the encoder as an additional input channel, similar to~\cite{perazzi2017learning}.
    Furthermore, we use a separate reference encoder to process the input to the matching branch.
    We highlight that the functions approximated by these two encoders differ, as the inputs to the  \textit{Reference Encoder} are aligned in time, but this is not the case for the \textit{Encoder} network.
    Finally, the hidden state of the RNN ($h_{t}$) is combined with
    the encoded features from the matching branch via a merge layer and passed to the decoder to predict the segmentation mask.
    The skip connections between the encoder and the decoder networks are not shown for simplicity.}
    \label{fig:architecture}
\end{figure*}

The S2S model is composed of an encoder-decoder architecture with a memory module in the bottleneck to memorize the target object and obtain temporal consistency in the predicted segmentation masks.
The overall design of this method is illustrated in \autoref{fig:architecture}.
In this model, the object masks are computed as in~\cite{xu2018youtub}:
\begin{equation}
    h_{0}, c_{0} = \text{Initializer}(x_{0}, y_{0})
\label{eq:1}
\end{equation}
\begin{equation}
    \Tilde{x}_{t}= \text{Encoder}(x_{t})
\label{eq:2}
\end{equation}
\begin{equation}
    h_{t}, c_{t} = \text{RNN}(\Tilde{x}_{t}, h_{t-1}, c_{t-1})
\label{eq:3}
\end{equation}
\begin{equation}
    \hat{y}_{t} = \text{Decoder}(h_{t})
\label{eq:4}
\end{equation}
where $x$ and $y$ refer to the RGB input image and the binary mask of the target object in the first frame.

One of the main limitations of RNN-based models, such as S2S, is the fixed-sized memory, which can be insufficient to capture the whole sequence and long-term dependencies~\cite{bahdanau2014neural}.
Therefore, as the sequence length grows, access to information from earlier time steps decreases.
This issue, together with the vanishing gradient problem, adversely impacts the segmentation quality in longer sequences.
This problem is especially critical in sequences with occlusion, where the object of interest can be absent for an extended period.

Another obstacle with this category of approaches is drift and error propagation.
Due to the recurrent connection, the model output is fed back to the network; 
as a result, the prediction error propagates to the future, and erroneous model predictions affect the performance for future time steps. 
This issue is another contributing factor to the performance drop in later frames.

\vspace{+1mm}
\textbf{Hybrid Mask Propagation}.
Based on the challenges in the RNN-based models, we propose a hybrid architecture, combining the RNN output with information derived from correspondence matching.
In our model, the segmentation mask is predicted using the location prior obtained from the RNN, as well as similarity matching between the video frames at $t-1$ and $t$.
Our intuition is that the merits of using the spatio-temporal model from RNN-based models and the matching-based methods are complementary.
In situations where multiple similar objects are present in the scene, the matching-based approaches struggle to distinguish between the different instances.
Hence, the location prior provided by the spatio-temporal features from the  RNN can resolve this ambiguity.
Moreover, the information obtained from similarity matching provides a reliable measurement for propagating the segmentation mask to the next time step (as investigated in~\cite{yang2019anchor} for zero-shot VOS). Using this additional data helps the model reduce the prediction error, improving the drift problem, and obtaining better segmentation quality for longer sequences.

To encode the frame at $t-1$, we redefine the initializer network's task in S2S to a reference encoder (as shown in \autoref{fig:architecture}), initializing the hidden states of the RNN module with $zeros$.
In our experiments, we observed that the initializer network does not play a crucial role, and it is possible to replace it with zero-initialization with little change in the performance.

To perform the similarity matching between the RNN hidden state ($h_{t}$) and the reference encoder's output features, one can use different techniques such as using the cosine distance between the feature vectors.
Here, we follow the design in~\cite{wug2018fast} and use a Global Convolution~\cite{peng2017large} to accomplish the task~(merge layer in \autoref{fig:architecture}).
Global Convolution~(GC) approximates a large kernel convolution layer efficiently with less number of parameters. 
The large kernel size is essential to model both the local connections (as required for localization) and the dense global connections required for accurate classification (foreground, background).
This way, the model directly accesses the features from time steps $0$ and $t-1$.
We note that this operation can also be interpreted as self-attention; as,
the features at the current time step, which share a higher similarity to the object features from the reference frames, get a higher weight via the convolution operation in the merge layer.

As shown in \autoref{fig:architecture}, we do not use weight sharing between the Reference Encoder and the Encoder, as we observed a considerable performance drop in doing so.
We believe the underlying reason is that the functions approximated by these two modules are different;
the inputs to the Reference Encoder are aligned in time while the inputs to the Encoder are not.
We highlight that compared to S2S~\cite{xu2018youtub}, the only added element is the light-weight Merge Layer~(\autoref{fig:architecture}).
The rest of the components remain unchanged, by modifying the task of the Initializer Network to Reference Encoder.

\textbf{Attention to the First Frame}.
As suggested in~\cite{ebert2017self} for the Video Prediction task, the first frame of the sequence is of significant importance as it contains the reference information which can be utilized for recovering from occlusion.
We note that by definition, the target object is present in the first frame.
By computing the correspondences between the object appearance after occlusion and in the first frame, the model is able to re-detect the target.
Additionally, \cite{yang2019anchor,wug2018fast} demonstrate the effectiveness of using the first frame as an anchor or reference frame.
In~\cite{wug2018fast}, the authors propose a Siamese architecture that learns to segment the object of interest by finding the feature correspondences between the target object in the first frame and the current frame.
Although this model's performance suffers in scenarios with drastic appearance change, it reveals the importance of rigorously using the data in the first frame.
We use the same reference encoder and merge layer for integrating the first frame features.
We hypothesize that this modification can be considered as an attention mechanism~\cite{bahdanau2014neural}, where the attention span is limited to the first frame. 
Using attention is a standard solution to address this finite memory in the RNNs, by providing additional context to the memory module.
The context vector is usually generated from a weighted combination of the embeddings from all the time steps.
However, in high-dimensional data such as video, it would be computationally demanding to store the features and compute all the frames' attention weights.

The resulting architecture is shown in~\autoref{fig:architecture} and can be formulated as:
%jacc loss
\begin{equation}
% TODO: 0 part of Rd
    h_{0}, c_{0} = \bm{0}
\label{eq:5}
\end{equation}
\begin{equation}
    \Tilde{x}_{0} = \text{Reference\_Encoder}(x_0, y_0)
\label{eq:x0}
\end{equation}
\begin{equation}
    \Tilde{x}_{t-1} = \text{Reference\_Encoder}(x_{t-1}, y_{t-1})
\label{eq:xt_1}
\end{equation}
\begin{equation}
    \Tilde{x}_{t} = \text{Encoder}(x_{t}, y_{t-1})
\label{eq:6}
\end{equation}
\begin{equation}
    h_{t}, c_{t} = \text{RNN}(\Tilde{x}_{t}, h_{t-1}, c_{t-1})
\label{eq:7}
\end{equation}
\begin{equation}
    \hat{y} = \text{Decoder}(\Tilde{x}_{0}, \Tilde{x}_{t-1}, h)   
\label{eq:8}
\end{equation}
where \textit{x} and \textit{y} are the RGB image and the binary segmentation mask, and $\bm{0} \in R^{d}$ with $d$ as the feature dimension.
Here the merge layer is considered as part of the decoder.

\vspace{+1mm}
\textbf{Training Objective}.
For the loss function, we utilize a linear combination of the balanced Binary Cross-Entropy~(BCE) loss and an auxiliary loss~\cite{azimi2020revisiting}:
\begin{equation}
 L_\text{total} = \lambda\ L_\text{seg} + (1-\lambda)\ L_\text{aux}   
 \label{eq:total}
\end{equation}
The auxiliary task employed here is border classification. 
For this task, a border class is assigned to each pixel based on its location with respect to the object boundary, where the boundary target classes are assigned based on a distance transform~\cite{hayder2017boundary}.
This term provides fine-grained location information for each pixel resulting in improved boundary detection $F$-score.
For more details, please refer to~\cite{azimi2020revisiting}.

The balanced BCE loss is computed as in~\cite{caelles2017one} :
\begin{equation}
  \begin{aligned} 
    L_\text{seg}(\textbf{W}) & = \sum_{t=1}^{T}(-\beta\sum\limits_{j\in Y_{+}}\log P(y_{j}=1|X;\textbf{W})\\
      & -(1-\beta)\sum\limits_{j\in Y_{-}}\log P(y_{j}=0|X;\textbf{W}))
  \end{aligned}
  \label{eq:seg_loss}
\end{equation}
with $X$ as input, $\textbf{W}$ as the model parameters, $Y_{+}$ and $Y_{-}$ standing for the foreground and background groundtruth labels, $\beta=|Y_{-}|/|Y|$, and $|Y|=|Y_{-}|+|Y_{+}|$.
This loss addresses the data imbalance between the foreground and the background classes by the weighting factors $\beta$.
%%%%%%%%%%%%%%%%%%%%%%%%%%%%%%%%%%%%%%%%%%%%%%%%%%%%
%
%       
%       Implementation Details    
%
%
%%%%%%%%%%%%%%%%%%%%%%%%%%%%%%%%%%%%%%%%%%%%%%%%%%%%
\section{Implementation Details}
In this section, we explain the implementation details of our hybrid model.
The code and the pre-trained models will be made publicly available  \footnote{\url{https://github.com/fatemehazimi990/HS2S}}.
\subsection{Encoder Networks}
In the S2S model, a VGG network~\cite{simonyan2014very} is used as the backbone for the initializer and encoder networks.
In this work, we utilize a ResNet50~\cite{he2016deep} architecture, pre-trained on ImageNet~\cite{deng2009imagenet}.
We remove the last average pooling and the fully connected layers, which are specific for image classification.
Furthermore, we add an extra $1\times1$ convolution layer to reduce the number of output channels from $2048$ to $1024$.
% As in ResNet50, the spatial size is reduced by a factor of 4 in the first block, we embed and additional convolution layer with $64$ filters between the two spatial size reduction operations in order to utilize this added spatial level in the skip connections. 
%
The number of input channels is altered to $4$, as we feed the RGB image and the binary segmentation mask to the encoder.
We utilize skip connections~\cite{ronneberger2015u} between the encoder and the decoder at every spatial resolution of the feature map ($5$ levels in total) to capture the fine details lost in the pooling operations and reducing the spatial size of the feature map.
Moreover, we use an additional RNN module in the first skip connection, as suggested in~\cite{azimi2020revisiting}.
The impact of changing the backbone network in the S2S model from VGG to ResNet on the segmentation accuracy is studied in \autoref{tab:component}.
\subsection{RNN and Merge Layer}
For the RNN component, we use a ConvLSTM layer~\cite{xingjian2015convolutional}, with a kernel-size of $3\times3$ and $1024$ filters.
As suggested in~\cite{xu2018youtub}, $Sigmoid$ and $ReLU$ activations are used for the gate and state outputs, respectively.

The merging layer's role is to perform correspondence matching between the RNN hidden state (the spatio-temporal features) and the outputs from the reference encoder.
There are different ways that can be used for this layer based on similarity matching and cosine distance.
Similar to~\cite{wug2018fast}, we utilize Global Convolution (GC) layers~\cite{peng2017large} for this function.
Two GC layers with an effective kernel size of $7\times7$ are employed to combine the RNN hidden state with the reference features and the features from the previous time step ($\Tilde{x}_{0}$ and $\Tilde{x}_{t-1}$ as in \Cref{eq:x0,eq:xt_1}).
The output of these two layers are then merged using a $1\times1$ convolution and then fed into the decoder network.
\subsection{Decoder}
The decoder network consists of five up-sampling layers followed by $5\times5$ convolution layers with $512$, $256$, $128$, $64$, and $64$ number of filters, respectively.
In the last layer, a $Conv_{1\times1}$ maps the $64$ channels to $1$ and a $Sigmoid$ activation is used to generate the binary segmentation scores (for the foreground and background classes).
The features from the skip connections are merged into the decoder using a $1\times1$ convolution layers.
$ReLU$ activation is used after each convolution layer, except for the last layer, where we use $Sigmoid$ activation to generate the segmentation output.
\subsection{Training Details}
For data augmentation, we apply random horizontal flipping as well as affine transformations.
The $\lambda$ in \autoref{eq:total} is set to $0.8$.
We use Adam optimizer~\cite{kingma2014adam} with an initial learning rate of $10^{-4}$.
We gradually lower the learning rate in the final phase of training when the loss is stable.
During the training, we use video snippets with $5$ to $10$ frames and a batch size of $16$. 

Additionally, we apply a curriculum learning method as suggested for sequence prediction tasks~\cite{bengio2015scheduled}.
To this end, we use the ground-truth for the segmentation mask input in the earlier stages of training where the model output is not yet satisfactory. 
This phase is known as \textit{teacher forcing}.
Next, with a pre-defined probabilistic scheme~\cite{bengio2015scheduled}, we randomly choose between using the ground-truth or the model-generated segmentation mask, on a per-frame basis.
This process helps to close the gap between the training and inference data distributions (during the inference, only the model-generated masks are used).
%
%%%%%%%%%%%%%%%%%%%%%%%%%%%%%%%%%%%%%%%%%%%%%%%%%%%%
%
%       
%       Experiments    
%
%
%%%%%%%%%%%%%%%%%%%%%%%%%%%%%%%%%%%%%%%%%%%%%%%%%%%%
\section{Experiments}
This section provides the experimental results for our hybrid model and a comparison with other state-of-the-art methods.
Additionally, we analyze our hybrid model's behavior on the two challenging scenarios occlusion and long sequences.
\subsection{Evaluation on Youtube-VOS and DAVIS2017}
We evaluate our model on the Youtube-VOS~\cite{xu2018youtub} dataset (the largest for Video Object Segmentation), as well as the DAVIS2017 dataset~\cite{pont20172017}.

We report the standard metrics of the task, namely \textit{Region Similarity} and \textit{Boundary Accuracy} ($F\&J$)~\cite{perazzi2016benchmark}.
The $F$ score measures the quality of the estimated segmentation boundaries and the Jaccard index $J$ measures the intersection over union area between the model output and the ground-truth segmentation mask.

\autoref{tab:yvos} shows a comparison of our model with other state-of-the-art methods.
The upper and lower sections include the methods with and without online training.
During the online training, the model is further fine-tuned on the first frame (where the segmentation mask is available) at test time;
Although this stage significantly improves the segmentation accuracy, it results in slow inference which is not practical for real-time applications.
Despite this, we see that our model without online training still outperforms the S2S model with online training.
The performance improvement compared to RGMP~\cite{wug2018fast}~(matching-based) and S2S~\cite{xu2018youtub}~(RNN-based) models strongly indicates that both propagation and matching information are required for better segmentation quality.
Moreover, our method achieves similar performance to STM~\cite{oh2019video} when training on the same amount of data~(not using synthetic data generated from image segmentation datasets) without relying on an external memory module;
Therefore, our model is less memory-constrained during the inference stage compared to methods using external memory that are prone to memory overflow for longer sequences~(in ~\cite{oh2019video}, authors save every 5th frame to the memory to avoid the GPU memory overflow during the test phase).

\begin{figure*}[t!]
    \centering
    \includegraphics[width=\linewidth]{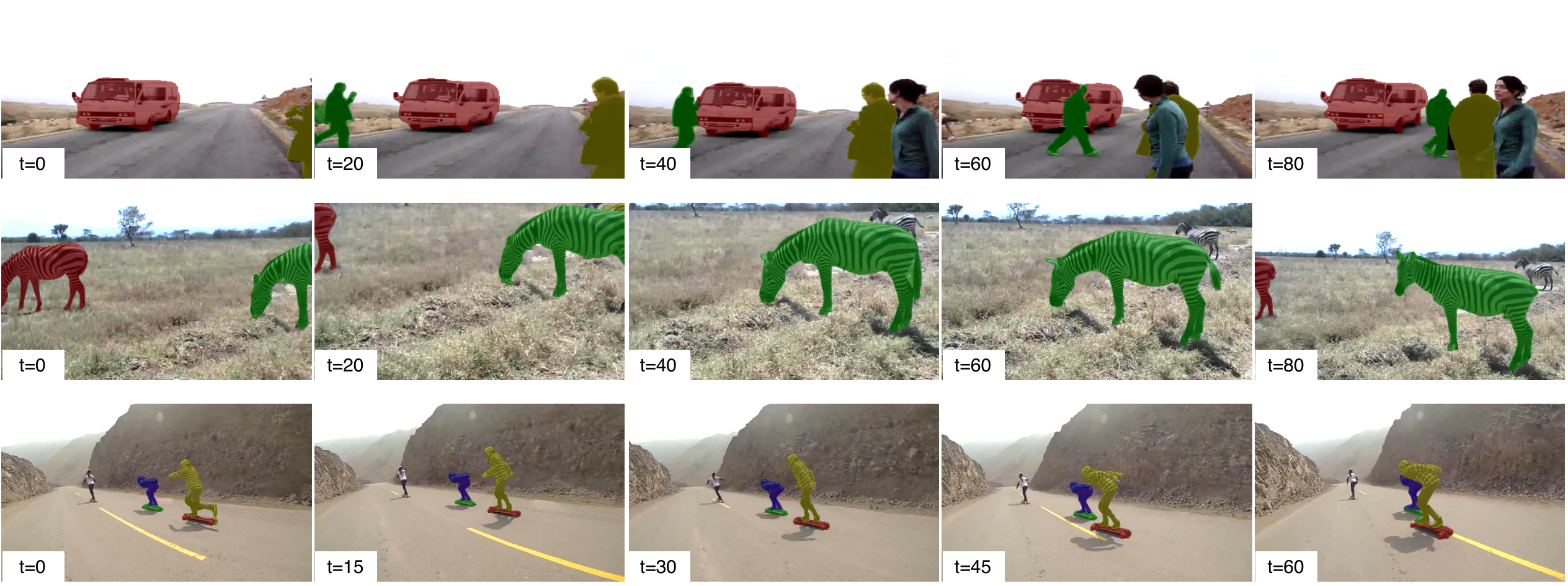}
    \caption{Visual samples of our model on Youtube-VOS validation set. As can be observed, our method can successfully segment sequences with similar object instances, even in the presence of occlusion.}
    \label{fig:sample1}
\end{figure*}

\autoref{fig:sample1} illustrates some visual examples from our model.
As we see, our model can properly track the target object in the presence of similar object instances as well as occlusion.
More visual samples are provided in the supplementary material.
\begin{table}
\caption{A comparison with the state-of-the-art methods on the Youtube-VOS dataset~\cite{xu2018youtub}. The upper part of the table shows models with online training, the lower part without.
All scores are in percent.
RVOS, S2S, and S2S++ are the RNN-based architectures.
As shown in this table, our hybrid model outperforms the S2S(no-OL) baseline model with an average improvement of 11.2 pp.
STM- refers to results in~\cite{oh2019video}, with the same amount of training data for a fair comparison. 
We can see that our method can achieve similar results to STM-, without requiring an external memory module. 
}
\centering
\begin{tabular}{|c c c c| }
\hline
Method & $\mathit{J}$ & $\mathit{F}$ & $\mathit{F\&J}$\\
\hline\hline
OSVOS~\cite{Man+18b} & 57.0 & 60.6 & 58.8\\
MaskTrack~\cite{perazzi2017learning} & 52.5 & 53.7 & 50.6\\
%OnAVOS~\cite{voigtlaender2017online}  & 53.4 & 57.1\\
S2S(OL)~\cite{xu2018youtub} & \textbf{63.25} & \textbf{65.6} & \textbf{64.4}\\
\hline
OSMN~\cite{yang2018efficient} & 50.3 & 52.1 & 51.2 \\
RGMP~\cite{wug2018fast} & 52.4 & 55.3 & 53.8\\
RVOS~\cite{ventura2019rvos} & 54.6 & 59.1 & 56.8\\
A-GAME~\cite{johnander2019generative} & 64.3 & 67.9 & 66.1\\
S2S(no-OL)~\cite{xu2018youtub} & 57.5 & 57.9 & 57.7\\
S2S++~\cite{azimi2020revisiting} & 58.8 & 63.2 & 61.0\\
STM-~\cite{oh2019video} & - & - & 68.2 \\
TVOS~\cite{zhang2020transductive} & 65.4 & 70.5 & 67.2 \\
HS2S~(ours)& \textbf{66.1} & \textbf{71.7} & \textbf{68.9}\\
\hline
\end{tabular} 
\label{tab:yvos}
\end{table}

To assess the generalization of our model after training on Youtube-VOS, we freeze the weights and evaluate the model on DAVIS2017 dataset~\cite{pont20172017}.
The results can be seen in \autoref{tab:davis}.
We observe that our hybrid model outperforms the independent RNN-based and matching-based methods, even without fine-tuning on this dataset. 
% https://paperswithcode.com/task/semi-supervised-video-object-segmentation
\begin{table}
\caption{A comparison between the independent RNN-based~(RVOS) and matching-based~(RGMP) models and our hybrid method on the DAVIS2017 dataset~\cite{pont20172017}~(test-val).
HS2S- shows the results of our model trained on Youtube-VOS without fine-tuning on DAVIS2017.
The results of the S2S model on DAVIS2017 were not available.
}
\centering
\begin{tabular}{|c c c c|}
\hline
Method & $\mathit{J}$ & $\mathit{F}$ & $\mathit{F\&J}$\\
\hline\hline
S2S~\cite{xu2018youtub} & - & - & - \\
RVOS~\cite{ventura2019rvos} & 52.7 & 58.1 & 55.4\\
RGMP~\cite{wug2018fast} & 58.1 & 61.5 8 & 59.8\\
HS2S-~(ours) & \textbf{58.9} & \textbf{63.4} & \textbf{61.1} \\
\hline
\end{tabular}
\label{tab:davis}
\end{table}
\subsection{Analysis of Sequence Length and Occlusion}
To quantitatively assess our model's effectiveness, we evaluate it in challenging scenarios such as occlusion and longer sequences.
As the validation set of the Youtube-VOS dataset is not released, we use the 80:20-splits of the training set from~\cite{ventura2019rvos} for training and evaluation.
For the S2S model results, we further used our re-implementation as the code for their work is not publicly available.
Furthermore, we use the ResNet50 architecture as backbone for both models for a fair comparison (to our disadvantage, as it improves the overall evaluation score of $57.3\%$ for S2S (as reported in \cite{xu2018youtub}) to $60\%$ for our re-implementation S2S*).

\begin{figure}[t]
  \centering
  \includegraphics[width=1.\linewidth]{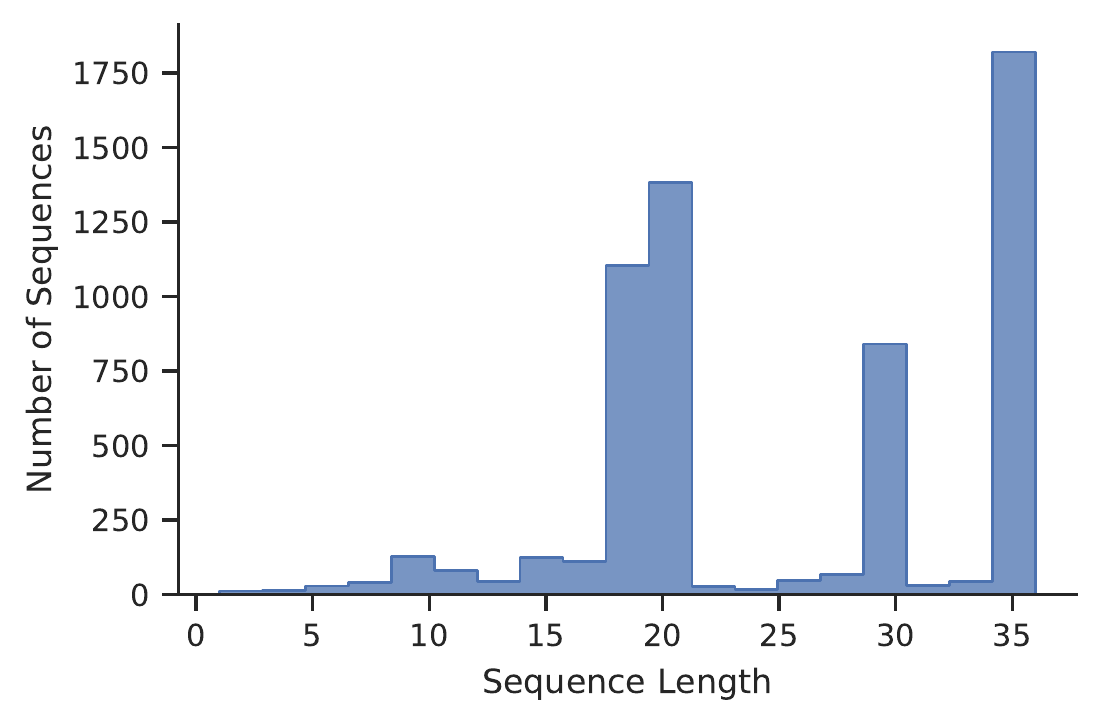}
   \caption{Distribution of the sequence length (per object) in the Youtube-VOS dataset.
   In Youtube-VOS, the video frame rate is reduced to $30$ fps, and the annotations are provided every fifth frame~($6$ fps).
   Therefore, a sequence with $36$ labeled frames spans $180$ time steps in the original frame rate.}
\label{fig:lenhistogram}
\end{figure}

\autoref{fig:lenhistogram} shows the sequence length distribution of the Youtube-VOS training set (one sequence per object in each video).
As can be seen, the length varies between $1$ to about $35$ frames in a very non-uniform fashion.
To study the impact of the video length on the segmentation scores, we pick the sequences longer than $20$ frames and measure the scores for frames with $t<10$ (considered as early frames) and frames with $t>20$ (considered as late frames), separately.
\begin{table}
\caption{A study on the impact of sequence length on the segmentation accuracy.
For this experiment, we picked the video sequences with more than $20$ frames.
Then we compute the $F$ and $J$ scores for frames earlier ($t<10$) and later ($t>20$) in the sequence.
As the results show, there is a performance drop as the time step increases.
However, our hybrid model's performance drops a lot less than the baseline's.}
\centering
\begin{tabular}{|l c c c c |}
\hline
Method & $\mathit{F_{l<10}}$ & $\mathit{J_{l<10}}$ & $\mathit{F_{l>20}}$ & $\mathit{J_{l>20}}$  \\
\hline\hline
S2S* & 74.4 & 73.7 & 54.5 & 54.6  \\
HS2S~(ours) & \textbf{77.1} & \textbf{76.3} & \textbf{65.5} & \textbf{64.2}  \\
\hline
\end{tabular}
\label{tab:len}
\end{table}
As presented in \autoref{tab:len}, we observe that the hybrid model improves the late frame accuracy significantly and reduces the performance gap between the early and late frames.
This observation confirms the effectiveness of the hybrid path for utilizing the information from spatio-temporal features as well as the correspondence matching.

\begin{figure}[t]
\centering
  \includegraphics[width=1.\linewidth]{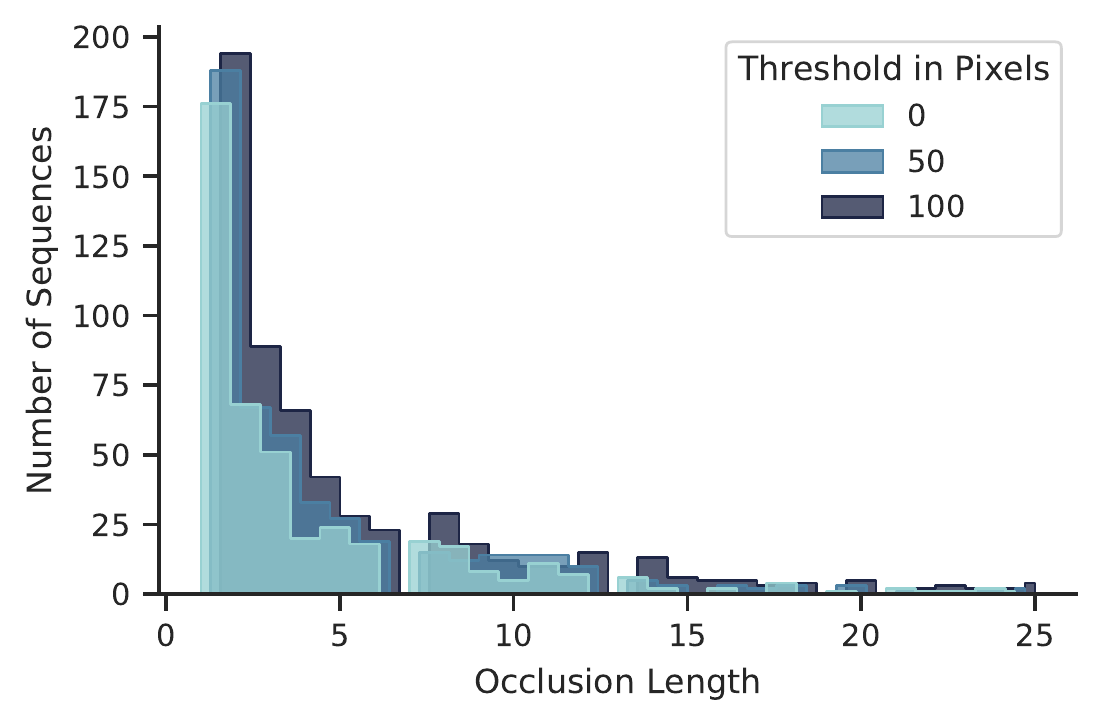}
   \caption{The number of occluded sequences (per object) in Youtube-VOS train set, for different occlusion lengths and with three occlusion thresholds (shifted by 1/3 for better visibility).}
\label{fig:occlusionhistogram}
\end{figure}

The histogram in \autoref{fig:occlusionhistogram} shows the number of sequences with occlusion in Youtube-VOS training set.
Each bin in the histogram shows the occlusion duration, and the $y$ axis indicates the number of sequences that belong to each bin.
As can be seen from this plot, the occlusion duration varies between $1$ to $25$ frames.
\begin{table}
\caption{A study on the impact of occlusion on the segmentation quality. 
The scores presented in this table are the average of $F$ and $J$ scores in percentages, when considering different thresholds (in pixels) for occlusion.
The $avg$ score refers to the average result for all the sequences in the 20-split.
For the other columns, we only considered the frames after ending the first occlusion period~(when the target object re-appears in the scene).
}
\centering
\begin{tabular}{|l c c c c|}
\hline
Method & $\mathit{avg}$ & $\mathit{th: 0}$ & $\mathit{th: 50}$ & $\mathit{th: 100}$ \\
\hline\hline
S2S* & 63.3 & 33.6 & 30.8 & 33.1\\
HS2S~(ours) & \textbf{69.0} & \textbf{40.2} & \textbf{39.3} & \textbf{47.7}\\
\hline
\end{tabular}
\label{tab:occlusion}
\end{table}
To study our model's effectiveness in handling occlusion, we report the scores for frames appearing \textit{after} a first occlusion in \autoref{tab:occlusion}.
An occlusion is considered a scenario where the object leaves the scene entirely and re-appears again.
As the areas below $100$ pixels are almost not visible (and could be considered as labeling noise), we also consider occlusions at three different thresholds of 0, 50, and 100 pixels.
As we can see in the table, occlusion is a challenging scenario with significantly lower scores than the average sequence scores.
However, our proposed approach again succeeds in defending its considerable improvement over the S2S baseline.
%%%%%%%%%%%%%%%%%%%%%%%%%%%%%%%%%%%%%%%%%%%%%%%%%%%%
%
%       
%       Ablation    
%
%
%%%%%%%%%%%%%%%%%%%%%%%%%%%%%%%%%%%%%%%%%%%%%%%%%%%%
\section{Ablation Study}
\label{ablation}
In this section, we present an ablation study on the impact of different components of our model.
In addition, we provide the results for a variant of our model where we use cosine similarity~\cite{wang2018non} for the merge layer instead of global convolution (referred to as HS2S$_{sim}$).

\autoref{tab:component} presents the segmentation scores when different components in our model are added one at a time.
The results for \textit{S2S*} are obtained from our re-implementation of the S2S model with ResNet50 backbone.
As it can be seen from the results, utilizing the first frame as the reference~(HS2S$_{0}$) and using the hybrid match-propagate strategy~(HS2S$_{t-1}$) both improve the segmentation quality.
Moreover, the enhancements add up when they are integrated into a single model~(HS2S).

\begin{table}
\caption{An ablation study on the impact of different components in our model.
S2S* is our re-implementation of the S2S method with the same backbone as our model, for a fair comparison (this version achieves a better segmentation accuracy).
S2S$_{0}$ refers to our model without the hybrid propagation, only using the first frame as reference.
S2S$_{t-1}$ is our model with hybrid propagation and without utilizing the first frame.
In HS2S$_\text{sim}$, we implemented the merge layer (\autoref{fig:architecture}) using cosine similarity instead of Global Convolution.}
\centering
\begin{tabular}{|c c c c|}
\hline
Method & $\mathit{J}$ & $\mathit{F}$ & $\mathit{F\&J}$\\
\hline\hline
S2S~\cite{xu2018youtub} & 57.5 & 57.9 & 57.7\\
S2S* & 59.1 & 63.7 & 61.4 \\
HS2S$_{0}$ & 64.0 & 68.95 & 66.5\\
HS2S$_{t-1}$ & 63.6 & 68.7 & 66.2\\
HS2S & \textbf{66.1} & \textbf{71.7} &\textbf{68.9} \\
HS2S$_{sim}$& 64.35 & 69.35 & 66.9\\
\hline
\end{tabular}
\label{tab:component}
\end{table}
\section{Conclusion}
In this work, we presented a hybrid architecture for the task of one-shot Video Object Segmentation. 
To this end, we combined the merits of RNN-based approaches and
models based on correspondence matching.
We showed that the advantages of these two categories are complementary, and can assist each other in challenging scenarios.
Our experiments demonstrate that both mechanisms are required for obtaining better segmentation quality.

Furthermore, we provided an analysis of two challenging scenarios: occlusion and longer sequences.
We observed that our hybrid model achieves a significant improvement in robustness compared to the baselines that rely on RNNs~\cite{xu2018youtub} and reference guidance~\cite{wug2018fast}.
However, occlusion remains an open challenge for future investigation, as the performance in this scenario is considerably lower than the average.
Moreover, we believe that integrating global information and modeling the interactions between the objects in the scene is a promising direction for future work.
% The unique characteristic of RNN is learning the motion model, which is not the case in other types of memory. RNN provides us the advantages of motion and location prior.
%

{\small
\bibliographystyle{ieee_fullname}
\bibliography{egbib}
}

\end{document}